\documentclass{article}
\usepackage{ijcai25}

\usepackage{times}
\usepackage{soul}
\usepackage{url}
\usepackage[hidelinks]{hyperref}
\usepackage[utf8]{inputenc}
\usepackage[small]{caption}
\usepackage{graphicx}
\usepackage{amsmath}
\usepackage{amsthm}
\usepackage{booktabs}
\usepackage{algorithm}
\usepackage{algorithmic}
\usepackage{multirow} 
\usepackage{amssymb}
\usepackage[switch]{lineno}

\title{GLDiTalker: Speech-Driven 3D Facial Animation with\\Graph Latent Diffusion Transformer}

\author{
Yihong Lin$^{1*}$\and
Zhaoxin Fan$^{2,3*}$\and
Xianjia Wu$^{4}$\and
Lingyu Xiong$^1$\and\\
Xiandong Li$^{4,5\dagger}$\and
Wenxiong Kang$^{1\dagger}$\and
Liang Peng$^4$\and
Songju Lei$^5$\And
Huang Xu$^4$
\affiliations
$^1$South China University of Technology\\
$^2$ Beijing Advanced Innovation Center for Future Blockchain and Privacy Computing, School of Artificial Intelligence, Beihang University\\
$^3$Hangzhou International Innovation Institute, Beihang University\\
$^4$Huawei Cloud\\
$^5$Nanjing University
\emails
lxdphys@smail.nju.edu.cn, 
auwxkang@scut.edu.cn
}

\begin{document}

\maketitle

\begin{abstract}
    Speech-driven talking head generation is a critical yet challenging task with applications in augmented reality and virtual human modeling. While recent approaches using autoregressive and diffusion-based models have achieved notable progress, they often suffer from modality inconsistencies, particularly misalignment between audio and mesh, leading to reduced motion diversity and lip-sync accuracy. To address this, we propose GLDiTalker, a novel speech-driven 3D facial animation model based on a Graph Latent Diffusion Transformer. GLDiTalker resolves modality misalignment by diffusing signals within a quantized spatiotemporal latent space. It employs a two-stage training pipeline: the Graph-Enhanced Quantized Space Learning Stage ensures lip-sync accuracy, while the Space-Time Powered Latent Diffusion Stage enhances motion diversity. Together, these stages enable GLDiTalker to generate realistic, temporally stable 3D facial animations. Extensive evaluations on standard benchmarks demonstrate that GLDiTalker outperforms existing methods, achieving superior results in both lip-sync accuracy and motion diversity.
\end{abstract}

\section{Introduction}

Speech-driven talking head generation aims to synthesize realistic and synchronized facial motion from speech, with applications in assistive technologies, such as virtual reality and augmented reality \cite{ping2013computer,wohlgenannt2020virtual}. Despite its potential, the task is challenging due to the many-to-many mapping between audio and facial motion, requiring a balance between accurate lip synchronization and natural expression generation.

\begin{figure}[t]
    \centering
    \includegraphics[scale=0.32]{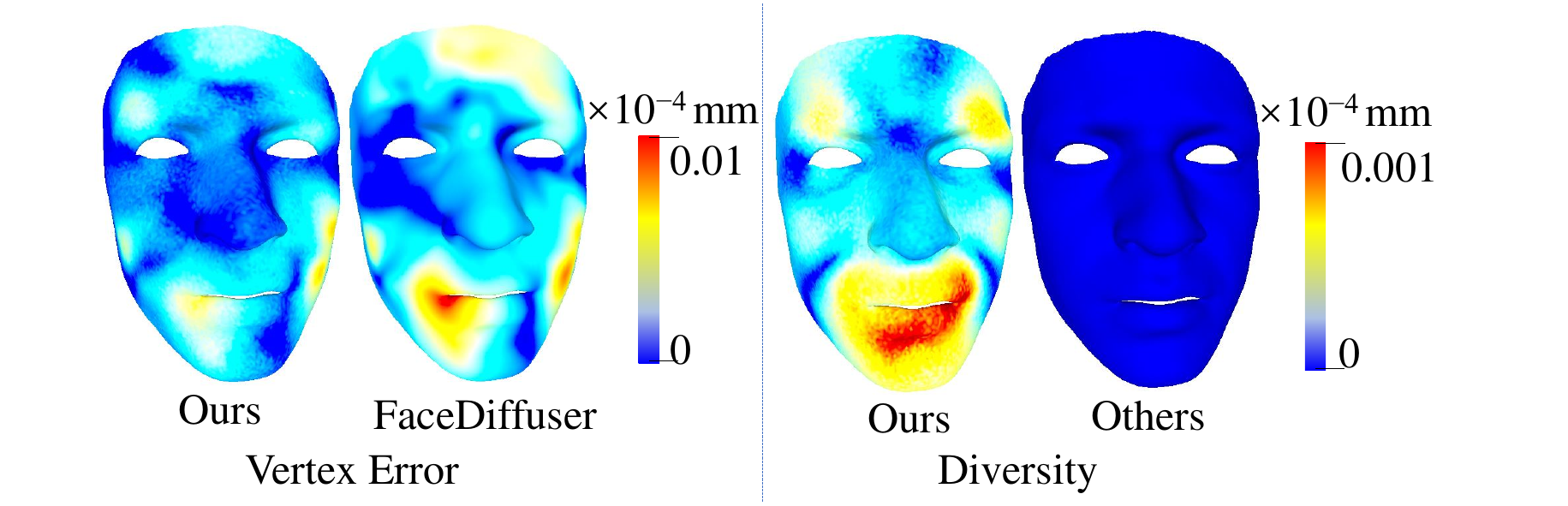}
    \caption{The left figures show the lip vertex error of GLDiTalker and FaceDiffuser and the right figures show the diversity of GLDiTalker and other methods, including FaceFormer, CodeTalker, TalkingStyle and FaceDiffuser. “Vertex Error” refers to the average $L_2$ loss between prediction and groundtruth. "Diversity" refers to the statistical mean of the motion differences between corresponding frames in multiple generation results within the same sequence. It is obvious that GLDiTalker has a significant advantage in both generation quality and motion diversity.}
    \label{fig1}
\end{figure}

To achieve realistic speech-driven talking head generation, various methods have been proposed.  Traditional approaches \cite{edwards2016jali,taylor2012dynamic,xu2013practical} rely on manual or rule-based designs, which are often costly and inflexible. In contrast, deep learning methods \cite{karras2017audio,peng2023selftalk} demonstrate stronger adaptability and have made significant progress. VOCA \cite{cudeiro2019capture} employs a CNN-based framework to generate cross-identity facial animations, while MeshTalk \cite{richard2021meshtalk} leverages U-Net to enhance expression diversity by modeling latent facial expression spaces. Transformer-based models such as FaceFormer \cite{fan2022faceformer} improve temporal and cross-modal alignment through mechanisms like Biased Cross-Modal Multi-Head Attention. Similarly, CodeTalker \cite{xing2023codetalker} introduces a VQ-VAE-based motion prior to reduce the uncertainty in audio-to-motion mapping. Despite these advancements, existing methods share common limitations. Most approaches rely on deterministic models, which generate identical 3D facial motions for the same input conditions. This contradicts real-world behavior, where non-verbal facial cues are inherently nondeterministic \cite{ng2022learning}. Even when the same individual repeats the same sentence, subtle variations in facial motion naturally occur, reflecting the complexity of human communication. By failing to capture this variability, current methods often suffer from modality inconsistencies, particularly misalignment between audio and mesh. These issues lead to reduced motion diversity and compromised lip-sync accuracy, ultimately undermining the realism and expressiveness of generated animations.

To tackle the issues, we propose GLDiTalker, a Graph Latent Diffusion Transformer for audio-driven talking head generation. GLDiTalker first encodes facial motion into a spatio-temporal quantized latent space, effectively addressing the audio-mesh modality misalignment to ensure accurate lip synchronization. It then diffuses the signal within this latent quantized spatio-temporal space, introducing a controlled degree of stochasticity to enhance motion diversity. Existing methods excel in either lip-sync accuracy or motion diversity but struggle with both due to audio-mesh modality misalignment. In contrast, GLDiTalker resolves this issue and achieves superior performance in both aspects, as shown in Fig. \ref{fig1}.

To achieve this goal, GLDiTalker adopts a carefully designed quantized space-time diffusion training pipeline, which consists of two complementary stages: the Graph Enhanced Quantized Space Learning Stage and the Space-Time Powered Latent Diffusion Stage.  In the first stage, GLDiTalker focuses on encoding facial motion into a spatio-temporal quantized latent space to ensure accurate lip synchronization. Unlike existing approaches that primarily emphasize temporal facial motion, our method simultaneously captures temporal correlations and spatial connectivities between vertices. To this end, we propose a novel Spatial Pyramidal SpiralConv Encoder, which extracts multi-scale features using spiral convolution across multiple levels. This design enables GLDiTalker to effectively integrate diverse levels of semantic information, laying a solid foundation for accurate and realistic lip synchronization. In the second stage, GLDiTalker employs a diffusion model to iteratively add and remove noise to the latent quantized facial motion features obtained from the first stage. This process introduces a controlled degree of stochasticity, enhancing motion diversity while maintaining the integrity of the learned features. Together, these two stages ensure that GLDiTalker achieves a balance between precise lip-sync and rich motion diversity, overcoming the limitations of previous methods.

Extensive experiments on existing datasets demonstrate that our method significantly outperforms previous state-of-the-art approaches in both lip-sync accuracy and motion diversity, effectively addressing the limitations of current techniques.  Our contribution can be summarized as:
\begin{itemize}
\item We introduce GLDiTalker, a novel framework that addresses the conflict between achieving accurate lip synchronization and diverse facial motion caused by audio-mesh modality misalignment.
\item GLDiTalker features a two-stage design: a Graph Enhanced Quantized Space Learning Stage for encoding spatio-temporal facial motion and a Space-Time Powered Latent Diffusion Stage for enhancing motion diversity.
\item Experiments on VOCASET and BIWI datasets demonstrate that GLDiTalker outperforms state-of-the-art methods in both lip-sync accuracy and motion diversity.
\end{itemize}

\section{Related Work}
\subsection{Speech-Driven 3D Facial Animation}
Speech-driven talking face generation has attracted significant attention and achieved remarkable progress in recent years \cite{karras2017audio,richard2021meshtalk,fan2022faceformer,xing2023codetalker,peng2023selftalk,wu2023speech,stan2023facediffuser,shen2025long}. Early methods, such as VOCA \cite{cudeiro2019capture}, introduce CNN-based frameworks for cross-identity face animation, while later works like Mimic \cite{fu2024mimic} and EmoTalk \cite{peng2023emotalk} focus on decoupling speech content, style, and emotion. Transformer-based approaches such as FaceFormer \cite{fan2022faceformer} improve audio-visual mapping through attention mechanisms, and CodeTalker \cite{xing2023codetalker} reduces cross-modal uncertainty using discrete motion priors. Recently, FaceDiffuser \cite{stan2023facediffuser} applies diffusion mechanisms to enhance the diversity of facial motion generation. Although these methods address various challenges, they still struggle to simultaneously achieve high accuracy and diversity due to limitations in modality alignment and inherent model constraints. To overcome these issues, we propose GLDiTalker, which introduces a quantized space-time diffusion training pipeline to better balance precision and diversity in speech-driven talking face generation.

\begin{figure*}[t]
    \centering
    \includegraphics[scale=0.66]{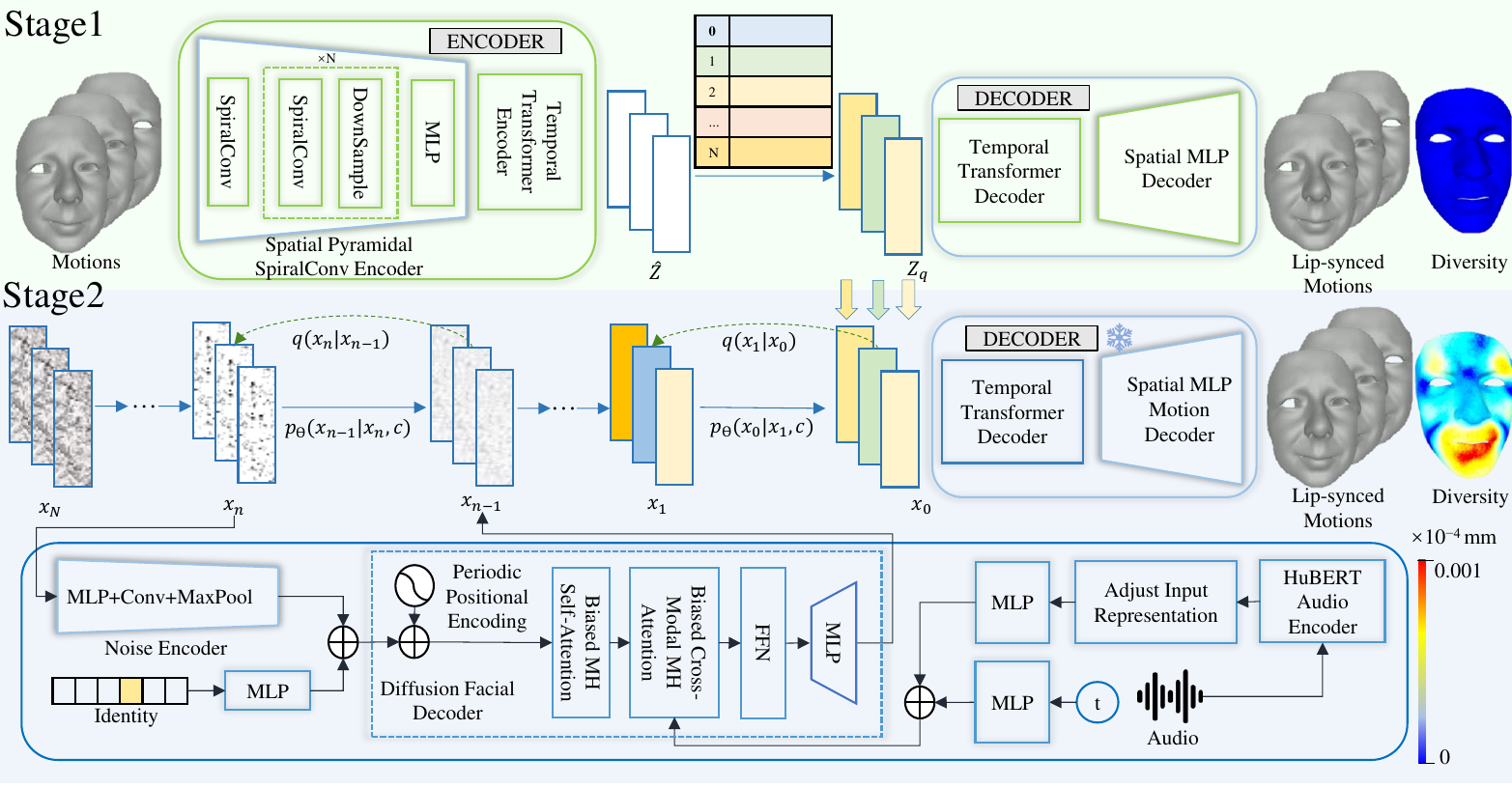}
    \caption{Overview of the proposed GLDiTalker. In order to enhance the lip-sync accuracy, Stage 1 employs the Spatial Pyramidal SpiralConv Encoder and temporal transformer encoder to sequentially process the input motions to obtain discrete latent facial motions. To enhance the motion diversity, Stage 2 then utilizes a diffusion-based iterative denoising technique to synthesize the latent facial motions from input speech and speaker identity, which is later decoded to a motion mesh sequence by the pretrained frozen decoder of the first stage. "$\oplus$" denotes addition and "N" denotes $N$ identical blocks, each consisting of a SpiralConv layer and a Downsample layer. The heatmaps show the statistical mean of the motion differences between corresponding frames in multiple generation results within the same sequence. }
    \label{fig2}
    
\end{figure*}

\subsection{Diffusion in Talking Head Animation}

Inspired by the diffusion processes in non-equilibrium thermodynamics, \cite{sohl2015deep} proposes diffusion, which differs from generative models like GAN \cite{goodfellow2020generative}, VAE \cite{kingma2013auto}, and Flow \cite{kingma2018glow}. Diffusion is a Markov process capable of generating high-quality images from Gaussian noise, exhibiting a certain degree of stochasticity \cite{shen2024imagpose,shen2025imagdressing}. Recently, many 2D talking head generation methods based on diffusion models have achieved highly realistic results \cite{park2022synctalkface,shen2022learning,zhang2023dream,shen2023difftalk,xu2024vasa}. For example, using audio as a condition, DREAM-Talk \cite{zhang2023dream} designs a two-stage diffusion-based framework that generates diverse expressions while maintaining accurate audio-lip synchronization. DiffTalk \cite{shen2023difftalk} further incorporates reference images and facial landmarks, enabling personality-aware synthesis and generalization across identities without fine-tuning. Similarly, VASA-1 \cite{xu2024vasa} employs diffusion in a latent space to model overall facial dynamics and head movements, effectively capturing subtle facial expressions and head motions. In the context of 3D facial animation, FaceDiffuser \cite{stan2023facediffuser} is the first to explicitly apply diffusion for speech-driven 3D facial motion generation. While diffusion methods inherently enhance diversity due to their stochastic nature, they often struggle to achieve high accuracy, particularly in terms of lip synchronization and motion precision. This trade-off between diversity and accuracy remains a significant challenge. To address this, we propose GLDiTalker, which leverages a quantized space-time diffusion training pipeline to balance diversity and accuracy.

\section{Method}

\subsection{Overview}

\subsubsection{Problem Definition}

The facial motion $M_{1:T}=(m_1, m_2, ..., m_T)$ is a sequence of vertex displacements $m_i \in \mathbb{R}^{V\times 3}$ applied to a neutral template face mesh $f \in \mathbb{R}^{V\times 3}$, which consists of $V$ vertices. The audio $A_{1:T}=(a_1, a_2, ..., a_T)$ is a sequence of speech snippets, where each $a_i \in \mathbb{R}^{C}$ contains $C$ samples aligned with the corresponding motion $m_i$. Our objective is to predict a mesh sequence $\hat M_{1:T}$ based on the audio $A_{1:T}$ and the talking style represented by one-hot vectors $S=(s_1, s_2, ..., s_I)$, where $I$ denotes the number of speaking styles. The final face animations are obtained by summing the neutral template and the predicted motion, expressed as $\hat F_{1:T}={\hat m_{1+f}, ..., \hat m_{T+f}}$. The output sequence $\hat F_{1:T}$ is expected to achieve accurate lip synchronization with the input audio while exhibiting diverse and expressive facial motions that reflect different speaking styles.

\subsubsection{Quantized Space-Time Diffusion Training Pipeline}
To solve the problem of failing to simultaneously achieve high lip-sync accuracy and motion diversity caused by modality misalignment, GLDiTalker follows a two-stage pipeline that consists of a Graph Enhanced Quantized Space Learning Stage and a Space-Time Powered Latent Diffusion Stage, as illustrated in Fig. \ref{fig2}. In the first stage, GLDiTalker focuses on quantization to enhance lip-sync accuracy. Specifically, it employs a spatio-temporal VQ-VAE \cite{van2017neural} based on graph convolution and transformer to model facial motion into a quantized facial motion prior, ensuring precise alignment between audio and facial motion. In the second stage, the focus shifts to diffusion for generating diverse motions. A transformer-based denoising network is adopted for the reverse diffusion process, which iteratively transforms a standard Gaussian distribution into the facial motion prior. This process is conditioned on audio, speaker identity, and the diffusion step, enabling the generation of realistic and expressive facial animations. The animation results can be directly obtained from the denoising outputs by using the decoder from the first stage.

Next, we introduce the Graph Enhanced Quantized Space Learning Stage and the Space-Time Powered Latent Diffusion Stage in detail, respectively.

\subsection{Stage 1: Graph Enhanced Quantized Space Learning for Lip-sync Accuracy Improvement}

Existing methods have limited results on lip-sync, mostly limited by the ability to model the explicit space or the accumulation of errors due to autoregression. To improve the robustness to cross-modal uncertainty, GLDiTalker proposes a Graph Enhanced Quantized Space Learning Stage inspired by VQ-VAE to model the facial motion into a quantized motion prior, as illustrated in the first stage in Fig. \ref{fig2}. A graph based encoder is trained with both the temporal information dependency and the spatial connectivity of mesh vertices taken into account. Specifically, spatial and temporal information are serially processed to get latent feature $\hat{Z}\in \mathbb{R}^{T \times H \times C}$ with a novel designed Spatial Pyramidal SpiralConv Encoder $E^V$ following by a temporal transformer encoder $E^T$:
\begin{eqnarray}\label{1}
{ E(M_{1:T})} = E^T(E^V(M_{1:T}))\rightarrow \hat Z_{1:T},
\end{eqnarray}
where $H$ and $C$ are the number of face components and channels, respectively. 

\begin{figure}[t]
    \centering
    \includegraphics[scale=0.35]{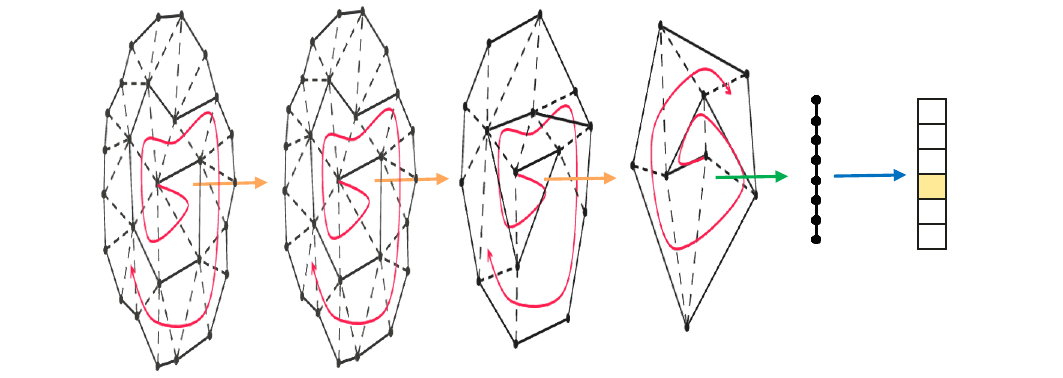}
    \caption{Architecture of Spatial Pyramidal SpiralConv Encoder. The spiral lines represent spiral convolution, the orange arrows represent downsampling, the green arrow represents flatten, and the blue arrow represents fusion by MLP.}
    \label{fig3}
\end{figure}

The Spatial Pyramidal SpiralConv Encoder consists of SpiralConv layers, downsampling layers and a fusion layer, as shown in Fig. \ref{fig3}. SpiralConv \cite{gong2019spiralnet++}, a graph-based convolution operator is introduced to process mesh data, which determines the convolution centre and produces a sequence of enumerated centre vertices based on adjacency, followed by the 1-ring vertices, the 2-ring vertices, and so on, until all vertices containing $k$ rings are included. 
SpiralConv determines the adjacency in the following way:

\begin{align}\label{2}
    0{-ring}(v) = {v}, \nonumber \\
    (k+1){-}ring(v) = \mathcal{NH}(k{-ring}(v)) \setminus k{-disk}(v), \nonumber \\
    k{-disk}(v) = \cup_{i=0,...,k}i{-ring}(v), 
\end{align}%
where $\mathcal{NH}(V)$ selects all vertices in the neighborhood of any vertex in set $V$. SpiralConv generates spiral sequences only once and explicitly encodes local information. It adopts a fully-connected layer for feature fusion:
\begin{align}\label{3} 
    SpiralConv(v) = W(f(k{-disk}(v)))+b,
\end{align}%
where $W$ and $b$ are learnable weights and bias. 

Then the SpiralConv layer and the downsampling layer are novelly composed into a Spiral Block, which reduces the number of vertices and increases the feature channels during feature extraction. After stacking multiple Spiral Blocks, the resulting graphs with a small scale are flattened and fed to the fusion layer (an MLP layer).

The Spatial Pyramidal SpiralConv Encoder is capable of extracting multi-scale features from the input graph by performing SpiralConv at multiple scales. This module enables the network to integrate semantic-level information at different levels. Lower-level pyramids can capture more detailed information, while higher-level pyramids focus more on abstract and semantic information. Pyramid network can effectively reduce the computational cost and improve the efficiency of the model in comparison to a single-scale network.

After getting the latent feature $\hat{Z}\in \mathbb{R}^{T\times C}$, the discrete facial motion latent sequence $Z_q\in \mathbb{R}^{T\times C}$ is obtained by an element-wise quantization function $Q(\cdot)$, which calculates by a nearest neighbour look-up in codebook $\mathcal{Z}=\left\{z_m\in \mathbb{R}^{C}\right\}_{m=1}^{M}$: 
\begin{eqnarray}\label{4}
{Z_q} = Q(\hat Z) \rightarrow \arg\min\ \left \| \hat z_{t} - z_{m}\right \|_2, z_{m} \in \mathcal{Z}.
\end{eqnarray}
Finally, temporal transformer decoder $D^T$ and spatial MLP decoder $D^V$ are adopted for self-reconstruction:
\begin{eqnarray}\label{5}
\hat{M}_{1:T} = D(Z_q) = D^V(D^T(Z_q)).
\end{eqnarray}

\subsubsection{Constraints}

In this stage, we train with the following loss functions:
\begin{eqnarray}\label{10}
L_{stage1} = \lambda_{rec1} L_{rec1} + \lambda_{quant} L_{quant},
\end{eqnarray}
where $\lambda_{rec1}=\lambda_{quant}=1$.

\emph{Motion Reconstruction Loss} calculates the L1 loss between the predicted animation sequence $\hat{M}_{1:T}$ and the reference facial motion $M_{1:T}$:
\begin{equation}\label{11}
{L}_{rec1} = \left \| \hat{M}_{1:T}-M_{1:T}\right \|_1.
\end{equation}

\emph{Quantization Loss} contains two intermediate code-level losses that update the codebook items by reducing the MSE distance between the codebook $\mathcal{Z}$ and latent features $\hat{Z}$:
\begin{equation}\label{12}
{L}_{quant} = \left \| sg(\hat{Z})-Z_q\right \|_2^2 + \beta \left \| \hat{Z}-sg(Z_q)\right \|_2^2,
\end{equation}
where $sg(\cdot)$ stands for the stop-gradient operator, which is defined as identity in the forward computation with zero partial derivatives; $\beta$ denotes a weighted hyperparameter, which is 0.25 in all our experiments.

\subsection{Stage 2: Space-Time Powered Latent Diffusion for Motion Diversity Enhancement}

To enhance the motion diversity that neglected by the deterministic models, GLDiTalker designs a Space-Time Powered Latent Diffusion Stage to introduce the quantized facial motion prior obtained from the first stage and subsequently train a speech-conditioned diffusion-based model on top of it. During the forward process, a Markov chain $q(x_n|x_{n-1})$ for $n \in \{1, ..., N\}$ gradually introduces Gaussian noise to the clean data sample $Z_q$ (denoted as $x_0$), resulting in a standard normal distribution $q(x_N|x_0)$:
\begin{eqnarray}\label{6}
q(x_N|x_0) = \prod \limits_{n=1}^N q(x_n|x_{n-1}), 
\end{eqnarray}
where $N$ is the diffusion step.

On the contrary, the distribution $p(x_{n-1}|x_n)$ is employed to reconstruct the clean data sample $x_0$ from the noise $x_N$ during the backward process. Conditioning on the audio $A_{1:T}$, one-hot speaker style embedding $s_i$ and diffusion step $n$, we apply a Markov chain $p(x_{n-1}|x_n,A_{1:T},s_i,n)$ to sequentially transform the distribution $p(x_N)$ into $p(x_0)$:
\begin{eqnarray}\label{7}
p(x_0|A_{1:T},s_i,N)&=& \prod \limits_{n=1}^N p(x_{n-1}|x_n,A_{1:T},s_i,n)\cdot\nonumber\\
&&p(x_N),
\end{eqnarray}
where $p(x_N) = \mathcal N(0,I)$. In practice, we use a denoising network $G$ that directly predicts the clean sample $\hat{x}_0$:
\begin{eqnarray}\label{8}
\hat{x}_0 = G(x_N,A_{1:T},s_i,N). 
\end{eqnarray}

As the second stage shown in Fig. \ref{fig2}, the architecture of the denoising network $G$ consists of five modules: (1) a noise encoder $E_{noise}$; (2) a style encoder $E_s$; (3) an audio encoder $E_a$; (4) a denoising step encoder $E_d$ and (5) a diffusion facial decoder $D$. 

Noise Encoder $E_{noise}$ reduces the dimension of the latent features, thereby retaining the most informative features while reducing the computational complexity of model training and inference, thus improving the efficiency and speed of the model.

Style Encoder $E_s$ encodes the one-hot speaker style embedding $s_i \in \mathbb{R}^{I} $ into a latent code.

Audio Encoder $E_a$ uses the released \emph{hubert-base-ls960} version of the HuBERT architecture pre-trained on 960 hours of 16kHz sampled speech audio. The feature extractor, feature projection layer and the initial two layers of the encoder are frozen, while the remaining parameters are set to be trainable, which enables HuBERT to better explore the motion information from audio. Since the facial motions might be captured with a frequency different from that of the speech tokens, we adjust input representation similar to \cite{stan2023facediffuser}. 

Denoising Step Encoder $E_d$ encodes the denoising step $n$ into a latent code.

Diffusion Facial Decoder $D$ is a transformer decoder that produces the final animation latent feature sample. The decoder process can be abstracted as follows:
\begin{equation}\label{9}
\begin{aligned}
\hat{x}_0&=D(E_{noise}(x_N),E_a(A_{1:T}),E_s(s_i),E_d(N)) \\
&= f(CA(SA(E_n(x_N)+E_s(s_i)),E_a(A_{1:T})+E_d(N))),
\end{aligned}
\end{equation}
where $CA(q,kv)$, $SA(\cdot)$ and $f(\cdot)$ mean Biased Cross-Modal Multi-Head Attention, Biased Causal Multi-Head Self-Attention and feed forward network, respectively; $q$ denotes query input features and $kv$ denotes key and value input features. The alignment mask \cite{fan2022faceformer} in $CA$ makes the motion features only attend to the speech features at the same position.

\subsubsection{Constraints}

The loss function of this stage consists of two items:
\begin{eqnarray}\label{13}
L_{stage2} = \lambda_{rec2} L_{rec2} + \lambda_{vel} L_{vel},
\end{eqnarray}
where $\lambda_{rec2}=\lambda_{vel}=1$.

Latent Feature Reconstruction Loss uses a Huber loss to supervise the recovered $\hat{x}_0$:
\begin{equation}\label{14} 
{L}_{rec2} = \left \| \hat{x}_0-x_0\right \|_H.
\end{equation}

Besides, Velocity Loss also employs a Huber loss to supervise inter-frame variations for smooth animation: 

\begin{equation}\label{15} 
{L}_{vel} = \left \| (\hat{x}_0^{2:T}-\hat{x}_0^{1:T-1})-(x_0^{2:T}-x_0^{1:T-1}))\right \|_H.
\end{equation}

\section{Experiments}

\subsection{Datasets and Implementations}
We conduct abundant experiments on two public 3D facial datasets, BIWI \cite{fanelli20103} and VOCASET \cite{cudeiro2019capture}, both of which have 4D face scans along with audio recordings. 

\noindent\textbf{BIWI dataset.} The BIWI dataset contains 40 sentences spoken by 14 subjects. The recordings are captured at 25 fps with 23370 vertices per mesh. We follow the data splits of the previous work \cite{fan2022faceformer} and only use the emotional data for fair comparisons. Specifically, the training set (BIWI-Train) contains 192 sentences, the validation set (BIWI-Val) contains 24 sentences, and the testing set are divided into two subsets, in which BIWI-Test-A contains 24 sentences spoken by 6 seen subjects during training and BIWI-Test-B contains 32 sentences spoken by 8 unseen subjects during training. BIWI-Test-A is used for both quantitative and qualitative evaluation and BIWI-Test-B is used for qualitative testing.

\noindent\textbf{VOCASET dataset.} The VOCASET dataset consists of 480 audio-visual pairs from 12 subjects. The facial motion sequences are captured at 60 fps with about 4 seconds in length. The 3D face meshes in VOCASET are registered to the FLAME topology \cite{FLAME:SiggraphAsia2017}, with 5023 vertices per mesh. Similar to \cite{fan2022faceformer}, we adopt the same training (VOCA-Train), validation (VOCA-Val) and testing (VOCA-Test) splits for qualitative testing.

\subsection{Quantitative Evaluation}
We adopt the following metrics for quantitative evaluation:

\begin{itemize}
\item\textbf{Lip vertex error (LVE)} measures the deviation of the lip vertices of the generated sequences relative to the ground truth by calculating the maximal L2 loss for each frame and averaging over all frames.

\item\textbf{Facial Dynamics Deviation (FDD)} measures the deviation of the upper face motion variation of the generated sequences relative to the ground truth. The standard deviation of the elements-wise L2 norm along the temporal axis at the upper face vertices of the generated sequence and the ground truth are initially calculated. Subsequently, their differences are solved and averaged.

\item\textbf{Diversity} measures the diversity of the generated facial motions from the same audio input. We calculate this metric across all samples in the BIWI-Test-A following \cite{ren2023diffusion}.

\end{itemize}

\begin{figure*}[t]
    \centering
    \includegraphics[scale=0.52]{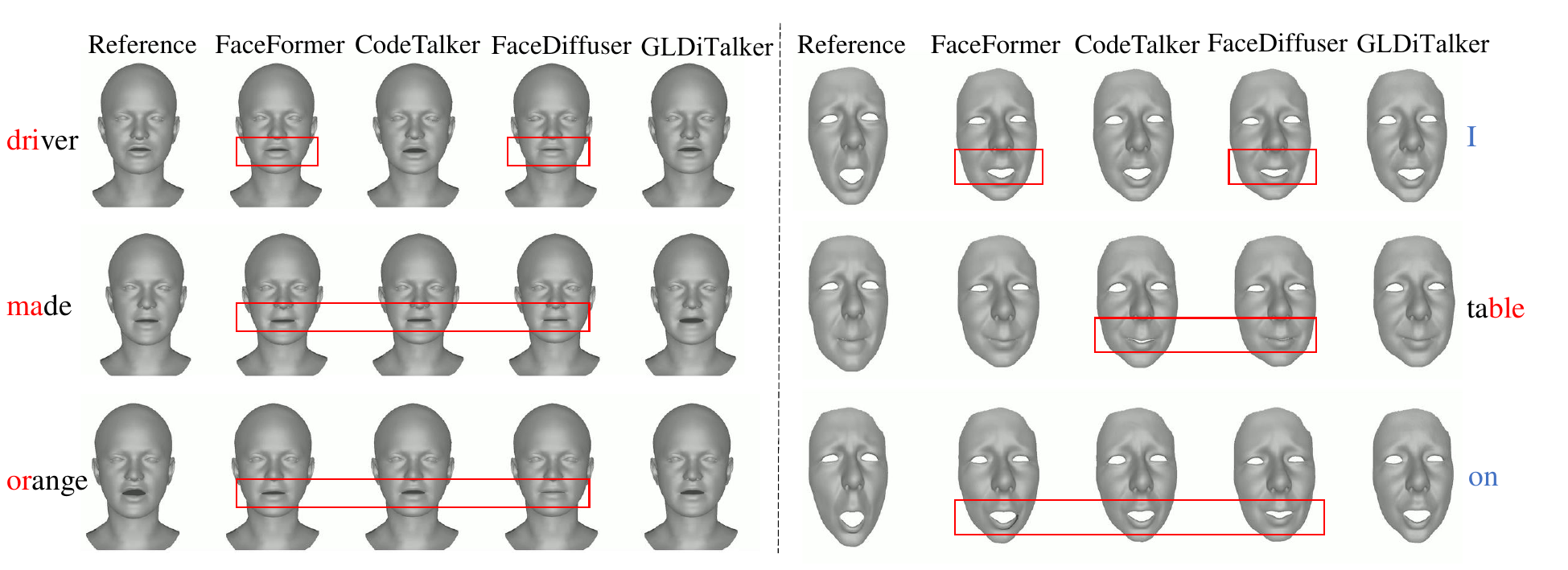}
    \caption{Qualitative Comparision on VOCASET-Test (left) and BIWI-Test-B (right). }
    \label{fig4}
\end{figure*}

\begin{figure}[t]
    \centering
    \includegraphics[scale=0.42]{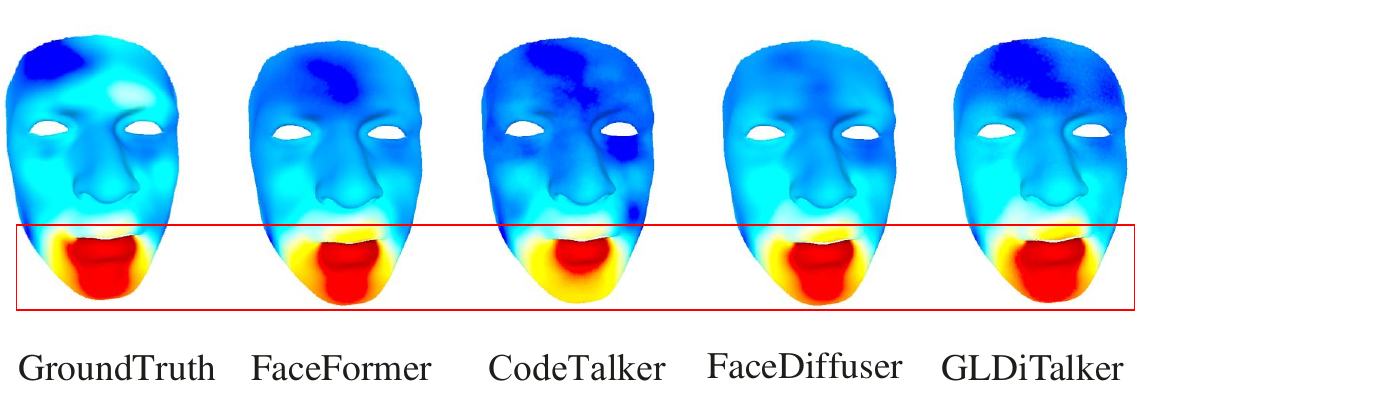}
    \caption{Standard deviation motion heatmaps of facial motion within a random sampled sequence, where dark blue means less motion variations and bright red means more motion variations.}
    \label{fig5}
\end{figure}

Since all subjects in BIWI-Test-A have all been seen in the training dataset, the model can learn the motion styles of the subjects in the testing set, which exhibits a relatively fixed output under speech-driven conditions. In contrast, the subjects in BIWI-Test-B and VOCASET-Test have not been included in the training dataset. Consequently, the model is only able to predict the animation based on the styles of the subjects in the training set. The results may differ significantly from the ground truth, but it may be reasonable to consider without the styles. Therefore, we only quantitatively compare the performance of our GLDiTalker with state-of-the-art methods on BIWI-Test-A, as in previous work \cite{xing2023codetalker} and \cite{stan2023facediffuser}. 

Quantitative evaluations are shown in Tab. \ref{table1}. Similar to the superior performance of CodeTalker compared to FaceFormer, GLDiTalker outperforms FaceDiffuser in terms of lip movement and upper facial expression motion prediction (0.1383$\times 10^{-4}$mm and 0.0751$\times 10^{-5}$mm decrease in LVE and FDD, respectively) by leveraging VQ-VAE to introduce a motion prior that enhances the robustness of the cross-modal mapping from audio to 3D vertex displacements. This strongly proves the effectiveness of our quantized space-time diffusion pipeline. Although TalkingStyle stands out with the lowest LVE and FDD, our approach ranks closely behind, showcasing only a minor difference. Furthermore, the highest Diversity (8.2176$\times 10^{-4}$mm) demonstrates the superiority of our model in capturing many-to-many mappings between audio and facial motion. The latent diffusion model that starts with random noise and generates latent features through multiple iterations of denoising extremely enhances motion diversity compared to deterministic methods and explicit diffusion model method.

\begin{table}[t]
    \centering
        \scalebox{0.75}{
        \begin{tabular}{@{}lccc@{}}
        \toprule
        Methods     & \begin{tabular}[c]{@{}c@{}}LVE $\downarrow$\\ ( $\times 10^{-4}$\text{mm})\end{tabular} & \begin{tabular}[c]{@{}c@{}}FDD $\downarrow$ \\ ( $\times 10^{-5}$\text{mm})\end{tabular} & \begin{tabular}[c]{@{}c@{}}Diversity $\uparrow$ \\ ($\times 10^{-4}$\text{mm}))\end{tabular}    \\ \midrule
        VOCA\shortcite{cudeiro2019capture}     & 6.5563     & 8.1816       & 0             \\
        MeshTalk\shortcite{richard2021meshtalk}     & 5.9181     & 5.1025       & 0             \\
        FaceFormer\shortcite{fan2022faceformer}      & 5.3077     & 4.6408       & 0             \\
        CodeTalker\shortcite{xing2023codetalker}      & 4.7914     & 4.1170       & 0          \\
        FaceDiffuser\shortcite{stan2023facediffuser}    & 4.7823     & 3.9225    &$5.6421\times10^{-5}$         \\   
        TalkingStyle\shortcite{talkingstyle} & 4.3646     & 3.8361    &0 \\ 
        \textbf{GLDiTalker}    & \textbf{4.6440}   & \textbf{3.8474}  & \textbf{8.2176}\\ 
        \hline

        \end{tabular}
}        
    \caption{Quantitative evaluations on BIWI-Test-A.}
    \label{table1}
\end{table}

\begin{table}[]
    \centering
        \scalebox{0.75}{
\begin{tabular}{@{}lcccc@{}}
\toprule
\multirow{2}{*}{Methods}      & \multicolumn{2}{c}{VOCASET-Test} & \multicolumn{2}{c}{BIWI-Test-B}    \\ \cmidrule(l){2-5} 
                             & \multicolumn{1}{l}{Realism$\uparrow$} & LipSync$\uparrow$ & Realism$\uparrow$       & LipSync$\uparrow$          \\ \midrule
FaceFormer         & 3.02         & 3.08    & 3.38     & 3.50          \\
CodeTalker         & 3.46         & 3.37   & 3.56     & 3.45          \\
FaceDiffuser       & 2.55         & 2.47   & 3.04     & 2.97          \\
\textbf{GLDiTalker}  & \textbf{3.94}   & \textbf{4.05}   & \textbf{3.85} & \textbf{4.00}   \\ \bottomrule
\end{tabular}
    }
    \caption{User study results.}
    \label{table2}
\end{table}

\subsection{Qualitative Evaluation}

We visually compare our GLDiTalker with FaceFormer, CodeTalker and FaceDiffuser and randomly assign the same speaking style as the conditional input to make a fair comparison. Fig. \ref{fig4} shows the animation results of the audio sequences from VOCASET-Test and BIWI-Test-B. We can observe that the lip movements produced by GLDiTalker are more accuracy. For vowel articulation, GLDiTalker exhibits a larger mouth opening than FaceFormer and FaceDiffuser, which is more consistent with GroundTruth. This can be observed in the mouth shape of the three examples in VOCASET-Test as well as 'I' and 'on' in BIWI-Test-B. For consonant articulation, GLDiTalker can fully close the mouth but CodeTalker and FaceDiffuser are less effective, which can be reflected by 'table' in BIWI-Test-B. Readers are recommended to watch the the Supplemental Video for more detail. In addition, we also visualize the standard deviation of facial movements of the four methods. Fig. \ref{fig5} shows that our GLDiTalker outperforms others in the range of facial motion variations, illustrating the generation diversity. In addition, the right side of Fig. \ref{fig1} demonstrates the highest motion diversity of GLDiTalker than the other methods. More comparisons are shown in \emph{Supplementary Video}.

\subsection{User Study}
In order to evaluate our model more comprehensively, we conduct a user study to evaluate realism and lip synchronization. Specifically, a questionnaire is designed for the purpose of comparing the effectiveness of FaceFormer, CodeTalker, FaceDiffuser and GLDiTalker with ground truth on 20 randomly selected samples from BIWI-Test-B and VOCASET-Test, respectively. Twenty participants are shown the questionnaire and asked to rate on a scale of 1-5. We count the Mean Opinion Score (MOS) of all methods. Tab. \ref{table2} shows that GLDiTalker gets the highest score in both two metrics, demonstrating its superiority on semantic comprehension and natural realism over other methods.

\subsection{Ablation Study}

In this section, we conduct an ablation study to the impact of our proposed quantized space-time diffusion training pipeline on the quality of generated 3D talking faces. All the experiments are conducted on BIWI dataset and Tab. \ref{table3} and Tab. \ref{table4} are the results from BIWI-Test-A. 

\begin{table}[]
    \centering
        \scalebox{0.75}{
        \begin{tabular}{@{}lccc@{}}
        \toprule
         Methods & \begin{tabular}[c]{@{}c@{}}LVE $\downarrow$ \\ ($\times 10^{-4}$\text{mm})\end{tabular}  & \begin{tabular}[c]{@{}c@{}}FDD $\downarrow$ \\ ($\times 10^{-5}$\text{mm})\end{tabular}  & \begin{tabular}[c]{@{}c@{}}Diversity $\uparrow$ \\ ($\times 10^{-4}$\text{mm})\end{tabular}  \\ \midrule
        \begin{tabular}[c]{@{}c@{}}w/o Graph Enhanced Quant-\\ized   Space Learning Stage\end{tabular}    & 4.7823   &  3.9225  &$5.6421\times10^{-5}$     \\  \midrule
        w/ Spatial MLP Encoder    & 4.8909   & 4.3384   &8.4251     \\ \midrule
        \begin{tabular}[c]{@{}c@{}}w/ Spatial Pyramidal \\ SpiralConv Encoder\end{tabular}     & \textbf{4.6440}     & \textbf{3.8474}  & \textbf{8.2176}       \\
        \hline

        \end{tabular}
}        
    \caption{Ablation study for Graph Enhanced Quantized Space Learning Stage on BIWI-Test-A.}
    \label{table3}
\end{table}

\begin{table}[]
    \centering
        \scalebox{0.75}{
        \begin{tabular}{@{}lcccc@{}}
        \toprule
        \begin{tabular}[c]{@{}c@{}}Kernel\\ Size\end{tabular}    & Dilation & Layer & \begin{tabular}[c]{@{}c@{}}LVE $\downarrow$ \\ ($\times 10^{-4}$\text{mm})\end{tabular}  & \begin{tabular}[c]{@{}c@{}}FDD $\downarrow$ \\ ($\times 10^{-5}$\text{mm})\end{tabular}  \\ \midrule
        \quad 5    & 1        & 4     & 4.9904      & 3.9252    \\
        \quad 9    & 1        & 2     & 4.9689      & 4.0887    \\
        \quad 9    & 1        & 4     & 4.7745      & 3.7400     \\
        \quad 9    & 2        & 3     & 5.1648      & 4.1967     \\       
        \quad 9    & 2        & 4     & \textbf{4.6440}     & \textbf{3.8474}  \\ 
        \hline

        \end{tabular}
}       
    \caption{Ablation study for hyperparameters of our Spatial Pyramidal SpiralConv Encoder on BIWI-Test-A.}
    \label{table4}
\end{table}

\begin{table}[t]
    \centering
        \scalebox{0.75}{
        \begin{tabular}{@{}ccc@{}}
        \toprule
        Methods    & \begin{tabular}[c]{@{}c@{}}LVE $\downarrow$ \\ ($\times 10^{-4}$\text{mm})\end{tabular}  & \begin{tabular}[c]{@{}c@{}}FDD $\downarrow$ \\ ($\times 10^{-5}$\text{mm})\end{tabular}  \\ \midrule
        \quad w/o Noise Encoder $E_{noise}$       & 5.5239      & 6.6215    \\
        \quad w Noise Encoder $E_{noise}$        & \textbf{4.6440}     & \textbf{3.8474}  \\ 
        \hline

        \end{tabular}
}      
    \caption{Ablation Study for Noise Encoder $E_{noise}$ on BIWI-Test-A.}
    \label{table5}
\end{table}

\subsubsection{Impact of Graph Enhanced Quantized Space Learning Stage}

Tab. \ref{table3} illustrates that our model achieves obvious superiority in lip-sync accuracy due to the Graph Enhanced Quantized Space Learning Stage with Spatial Pyramidal SpiralConv Encoder. The lowest LVE and FDD (4.6440$\times 10^{-4}$mm and 3.8474$\times 10^{-5}$mm in LVE and FDD, respectively) indicate the considerable impact on the overall visual quality of the generated faces and the comprehensibility of lip movements. With regard to the local connectivity of the graph, Spatial Pyramidal SpiralConv Encoder updates the representation of each mesh vertex by fusing the features with those of the neighbouring vertices. This process preserves the topology information of the vertices, thereby encouraging the model to effectively learn the representation of each vertex and the whole graph structure. In contrast, MLP is typically oriented towards regular input data with overly dense connections that do not take into account the structure of the graph.

A series of ablation experiments on kernel size, dilation coefficient, and layer numbers of Spatial Pyramidal SpiralConv Encoder are provided, as shown in Tab. \ref{table4}. With respect to kernel size, the data in the first and third rows indicates that a large kernel size confers a significant advantage in LVE and FDD (0.2159$\times 10^{-4}$mm and 0.1852$\times 10^{-5}$mm decrease in LVE and FDD, respectively). This may be attributed to the fact that larger convolution kernels have larger receptive field, thus enabling the integration of a greater quantity of information. Furthermore, in terms of the dilation coefficients, the third and fifth rows show that larger dilation coefficients lead to better lip synchronisation, although this is associated with a reduction in facial quality performance (0.1305$\times 10^{-4}$mm decrease in LVE and 0.1074$\times 10^{-5}$mm increase in FDD). We believe that the effect of lip synchronisation is more significant, so we retain the option of using a large dilation factor. Finally, for the number of layers, two sets of experiments are conducted to demonstrate the importance of the pyramid layer number. With a dilation factor of 1 (no dilation convolution operation), changing the number of layers from 4 to 2 leads to a decline in performance across all metrics (0.1944$\times 10^{-4}$mm and 0.3487$\times 10^{-5}$mm increase in LVE and FDD, respectively), as illustrated in the second and third rows; in the case of dilation factor of 2, modifying the number of layers from 4 to 3 also leads to worse performance (0.5208$\times 10^{-4}$mm and 0.3493$\times 10^{-5}$mm increase in LVE and FDD, respectively), as shown in the fourth and fifth rows. This suggests that an increase in the layer number can enhance feature extraction capability to comprehensively capture the intrinsic structure and regularity of the sparse vertex data effectively.

\subsubsection{Impact of Space-Time Powered Latent Diffusion Stage}
The last column of the Tab. \ref{table3} shows that using diffusion in the quantized space of Graph Enhanced Quantized Space Learning Stage leads to significant improvements in terms of motion diversity ($8.2176\times10^{-4}$mm). With the same input, the deterministic model can only produce a fixed output, which does not reflect the many-to-many correspondence between speech and facial motions in reality. Instead, our approach uses diffusion in the quantized space, which even has an advantage of several orders of magnitude over FaceDiffuser, which adopts diffusion in the explicit space. We also conducted experiments to remove the Noise Encoder $E_{noise}$, which resulted in a performance degradation, as shown in Tab. \ref{table5}, indicating the need of Noise Encoder.

\section{Conclusion and Discussion}

In conclusion, we propose GLDiTalker, a novel speech-driven 3D facial animation model that addresses the challenges of modality misalignment, lip-sync accuracy, and motion diversity in talking head generation. By introducing a two-stage pipeline, GLDiTalker ensures precise lip synchronization through the Graph-Enhanced Quantized Space Learning Stage and enhances motion diversity with the Space-Time Powered Latent Diffusion Stage. Extensive evaluations demonstrate that GLDiTalker achieves state-of-the-art performance, generating realistic, temporally stable 3D facial animations with both high lip-sync accuracy and diverse motion, making it a robust solution for speech-driven animation tasks.

\section*{Ethical Statement}
Face data can be used for generating content that may jeopardize privacy. We must act responsibly by considering the aspects related to privacy and ethics.

\section*{Acknowledgements}
This work was supported by the National Natural Science Foundation of China under Grant No. 62441617. It was also supported by the Beijing Natural Science Foundation under Grant No. 4254100, the Fundamental Research Funds for the Central Universities under Grant No. KG16336301, the China Postdoctoral Science Foundation under Grant No. 2024M764093, and by Beijing Advanced Innovation Center for Future Blockchain and Privacy Computing.

\section*{Contribution Statement}
Xiandong Li and Wenxiong Kang are the corresponding authors. They led the planning and execution of the research and provided financial support for the project leading to this publication. Yihong Lin and Zhaoxin Fan contributed equally to this work as co-first authors. Yihong Lin and Xiandong Li proposed the ideas. In addition, Yihong Lin designed the model and the implementation of the computer code, performed the experiments and wrote the initial draft. Zhaoxin Fan analyzed the study data and made critical review, commentary and revision including pre and post publication stages. Xianjia Wu and Lingyu Xiong assisted with figure preparation. Liang Peng, Songju Lei and Huang Xu proofread the manuscript. All authors have read and approved the final manuscript.

\appendix

\bibliographystyle{named}
\bibliography{glditalker}

@inproceedings{park2022synctalkface,
  title={Synctalkface: Talking face generation with precise lip-syncing via audio-lip memory},
  author={Park, Se Jin and Kim, Minsu and Hong, Joanna and Choi, Jeongsoo and Ro, Yong Man},
  booktitle={Proceedings of the AAAI Conference on Artificial Intelligence},
  volume={36},
  number={2},
  pages={2062--2070},
  year={2022}
}

@inproceedings{shen2022learning,
  title={Learning Dynamic Facial Radiance Fields for Few-Shot Talking Head Synthesis},
  author={Shen, Shuai and Li, Wanhua and Zhu, Zheng and Duan, Yueqi and Zhou, Jie and Lu, Jiwen},
  booktitle={Computer Vision--ECCV 2022: 17th European Conference, Tel Aviv, Israel, October 23--27, 2022, Proceedings, Part XII},
  pages={666--682},
  year={2022}
}

@article{edwards2016jali,
  title={Jali: an animator-centric viseme model for expressive lip synchronization},
  author={Edwards, Pif and Landreth, Chris and Fiume, Eugene and Singh, Karan},
  journal={ACM Transactions on graphics (TOG)},
  volume={35},
  number={4},
  pages={1--11},
  year={2016},
  publisher={ACM New York, NY, USA}
}

@inproceedings{taylor2012dynamic,
  title={Dynamic units of visual speech},
  author={Taylor, Sarah L and Mahler, Moshe and Theobald, Barry-John and Matthews, Iain},
  booktitle={Proceedings of the 11th ACM SIGGRAPH/Eurographics conference on Computer Animation},
  pages={275--284},
  year={2012}
}

@inproceedings{wu2023speech,
  title={Speech-Driven 3D Face Animation with Composite and Regional Facial Movements},
  author={Wu, Haozhe and Zhou, Songtao and Jia, Jia and Xing, Junliang and Wen, Qi and Wen, Xiang},
  booktitle={Proceedings of the 31st ACM International Conference on Multimedia},
  pages={6822--6830},
  year={2023}
}

@incollection{xu2013practical,
  title={A practical and configurable lip sync method for games},
  author={Xu, Yuyu and Feng, Andrew W and Marsella, Stacy and Shapiro, Ari},
  booktitle={Proceedings of Motion on Games},
  pages={131--140},
  year={2013}
}

@inproceedings{cudeiro2019capture,
  title={Capture, learning, and synthesis of 3D speaking styles},
  author={Cudeiro, Daniel and Bolkart, Timo and Laidlaw, Cassidy and Ranjan, Anurag and Black, Michael J},
  booktitle={Proceedings of the IEEE/CVF Conference on Computer Vision and Pattern Recognition},
  pages={10101--10111},
  year={2019}
}

@inproceedings{fan2022faceformer,
  title={Faceformer: Speech-driven 3d facial animation with transformers},
  author={Fan, Yingruo and Lin, Zhaojiang and Saito, Jun and Wang, Wenping and Komura, Taku},
  booktitle={Proceedings of the IEEE/CVF Conference on Computer Vision and Pattern Recognition},
  pages={18770--18780},
  year={2022}
}

@article{karras2017audio,
  title={Audio-driven facial animation by joint end-to-end learning of pose and emotion},
  author={Karras, Tero and Aila, Timo and Laine, Samuli and Herva, Antti and Lehtinen, Jaakko},
  journal={ACM Transactions on Graphics (TOG)},
  volume={36},
  number={4},
  pages={1--12},
  year={2017},
  publisher={ACM New York, NY, USA}
}

@InProceedings{peng2023emotalk,
    author    = {Peng, Ziqiao and Wu, Haoyu and Song, Zhenbo and Xu, Hao and Zhu, Xiangyu and He, Jun and Liu, Hongyan and Fan, Zhaoxin},
    title     = {EmoTalk: Speech-Driven Emotional Disentanglement for 3D Face Animation},
    booktitle = {Proceedings of the IEEE/CVF International Conference on Computer Vision (ICCV)},
    month     = {October},
    year      = {2023},
    pages     = {20687-20697}
}

@inproceedings{richard2021meshtalk,
  title={Meshtalk: 3d face animation from speech using cross-modality disentanglement},
  author={Richard, Alexander and Zollh{\"o}fer, Michael and Wen, Yandong and De la Torre, Fernando and Sheikh, Yaser},
  booktitle={Proceedings of the IEEE/CVF International Conference on Computer Vision},
  pages={1173--1182},
  year={2021}
}

@inproceedings{xing2023codetalker,
  title={Codetalker: Speech-driven 3d facial animation with discrete motion prior},
  author={Xing, Jinbo and Xia, Menghan and Zhang, Yuechen and Cun, Xiaodong and Wang, Jue and Wong, Tien-Tsin},
  booktitle={Proceedings of the IEEE/CVF Conference on Computer Vision and Pattern Recognition},
  pages={12780--12790},
  year={2023}
}

@article{van2017neural,
  title={Neural discrete representation learning},
  author={Van Den Oord, Aaron and Vinyals, Oriol and others},
  journal={Advances in neural information processing systems},
  volume={30},
  year={2017}
}

@inproceedings{peng2023selftalk,
  title={Selftalk: A self-supervised commutative training diagram to comprehend 3d talking faces},
  author={Peng, Ziqiao and Luo, Yihao and Shi, Yue and Xu, Hao and Zhu, Xiangyu and Liu, Hongyan and He, Jun and Fan, Zhaoxin},
  booktitle={Proceedings of the 31st ACM International Conference on Multimedia},
  pages={5292--5301},
  year={2023}
}

@inproceedings{stan2023facediffuser,
  title={Facediffuser: Speech-driven 3d facial animation synthesis using diffusion},
  author={Stan, Stefan and Haque, Kazi Injamamul and Yumak, Zerrin},
  booktitle={Proceedings of the 16th ACM SIGGRAPH Conference on Motion, Interaction and Games},
  pages={1--11},
  year={2023}
}

@inproceedings{gong2019spiralnet++,
  title={Spiralnet++: A fast and highly efficient mesh convolution operator},
  author={Gong, Shunwang and Chen, Lei and Bronstein, Michael and Zafeiriou, Stefanos},
  booktitle={Proceedings of the IEEE/CVF international conference on computer vision workshops},
  pages={0--0},
  year={2019}
}

@article{ping2013computer,
  title={Computer facial animation: A review},
  author={Ping, Heng Yu and Abdullah, Lili Nurliyana and Sulaiman, Puteri Suhaiza and Halin, Alfian Abdul},
  journal={International Journal of Computer Theory and Engineering},
  volume={5},
  number={4},
  pages={658},
  year={2013},
  publisher={IACSIT Press}
}

@article{wohlgenannt2020virtual,
  title={Virtual reality},
  author={Wohlgenannt, Isabell and Simons, Alexander and Stieglitz, Stefan},
  journal={Business \& Information Systems Engineering},
  volume={62},
  pages={455--461},
  year={2020},
  publisher={Springer}
}

@inproceedings{sohl2015deep,
  title={Deep unsupervised learning using nonequilibrium thermodynamics},
  author={Sohl-Dickstein, Jascha and Weiss, Eric and Maheswaranathan, Niru and Ganguli, Surya},
  booktitle={International conference on machine learning},
  pages={2256--2265},
  year={2015},
  organization={PMLR}
}

@inproceedings{
xu2024vasa,
title={{VASA}-1: Lifelike Audio-Driven Talking Faces Generated in Real Time},
author={Sicheng Xu and Guojun Chen and Yu-Xiao Guo and Jiaolong Yang and Chong Li and Zhenyu Zang and Yizhong Zhang and Xin Tong and Baining Guo},
booktitle={The Thirty-eighth Annual Conference on Neural Information Processing Systems},
year={2024},
}

@inproceedings{ren2023diffusion,
  title={Diffusion motion: Generate text-guided 3d human motion by diffusion model},
  author={Ren, Zhiyuan and Pan, Zhihong and Zhou, Xin and Kang, Le},
  booktitle={ICASSP 2023-2023 IEEE International Conference on Acoustics, Speech and Signal Processing (ICASSP)},
  pages={1--5},
  year={2023},
  organization={IEEE}
}

@inproceedings{shen2023difftalk,
  title={Difftalk: Crafting diffusion models for generalized audio-driven portraits animation},
  author={Shen, Shuai and Zhao, Wenliang and Meng, Zibin and Li, Wanhua and Zhu, Zheng and Zhou, Jie and Lu, Jiwen},
  booktitle={Proceedings of the IEEE/CVF Conference on Computer Vision and Pattern Recognition},
  pages={1982--1991},
  year={2023}
}

@article{zhang2023dream,
  title={Dream-talk: diffusion-based realistic emotional audio-driven method for single image talking face generation},
  author={Zhang, Chenxu and Wang, Chao and Zhang, Jianfeng and Xu, Hongyi and Song, Guoxian and Xie, You and Luo, Linjie and Tian, Yapeng and Guo, Xiaohu and Feng, Jiashi},
  journal={arXiv preprint arXiv:2312.13578},
  year={2023}
}

@inproceedings{fu2024mimic,
  title={Mimic: Speaking Style Disentanglement for Speech-Driven 3D Facial Animation},
  author={Fu, Hui and Wang, Zeqing and Gong, Ke and Wang, Keze and Chen, Tianshui and Li, Haojie and Zeng, Haifeng and Kang, Wenxiong},
  booktitle={Proceedings of the AAAI Conference on Artificial Intelligence},
  volume={38},
  number={2},
  pages={1770--1777},
  year={2024}
}

@article{goodfellow2020generative,
  title={Generative adversarial networks},
  author={Goodfellow, Ian and Pouget-Abadie, Jean and Mirza, Mehdi and Xu, Bing and Warde-Farley, David and Ozair, Sherjil and Courville, Aaron and Bengio, Yoshua},
  journal={Communications of the ACM},
  volume={63},
  number={11},
  pages={139--144},
  year={2020},
  publisher={ACM New York, NY, USA}
}

@article{kingma2013auto,
  title={Auto-encoding variational bayes},
  author={Kingma, Diederik P and Welling, Max},
  journal={International Conference on Learning Representations (ICLR)},
  year={2014}
}

@article{kingma2018glow,
  title={Glow: Generative flow with invertible 1x1 convolutions},
  author={Kingma, Durk P and Dhariwal, Prafulla},
  journal={Advances in neural information processing systems},
  volume={31},
  year={2018}
}

@article{fanelli20103,
  title={A 3-d audio-visual corpus of affective communication},
  author={Fanelli, Gabriele and Gall, Juergen and Romsdorfer, Harald and Weise, Thibaut and Van Gool, Luc},
  journal={IEEE Transactions on Multimedia},
  volume={12},
  number={6},
  pages={591--598},
  year={2010},
  publisher={IEEE}
}

@article{FLAME:SiggraphAsia2017,
  title = {Learning a model of facial shape and expression from {4D} scans},
  author = {Li, Tianye and Bolkart, Timo and Black, Michael. J. and Li, Hao and Romero, Javier},
  journal = {ACM Transactions on Graphics, (Proc. SIGGRAPH Asia)},
  volume = {36},
  number = {6},
  year = {2017},
  url = {https://doi.org/10.1145/3130800.3130813}
}

@inproceedings{ng2022learning,
  title={Learning to listen: Modeling non-deterministic dyadic facial motion},
  author={Ng, Evonne and Joo, Hanbyul and Hu, Liwen and Li, Hao and Darrell, Trevor and Kanazawa, Angjoo and Ginosar, Shiry},
  booktitle={Proceedings of the IEEE/CVF Conference on Computer Vision and Pattern Recognition},
  pages={20395--20405},
  year={2022}
}

@ARTICLE{talkingstyle,
  author={Song, Wenfeng and Wang, Xuan and Zheng, Shi and Li, Shuai and Hao, Aimin and Hou, Xia},
  journal={IEEE Transactions on Visualization and Computer Graphics}, 
  title={TalkingStyle: Personalized Speech-Driven 3D Facial Animation with Style Preservation}, 
  year={2024},
  pages={1-12},}

@article{shen2024imagpose,
  title={Imagpose: A unified conditional framework for pose-guided person generation},
  author={Shen, Fei and Tang, Jinhui},
  journal={Advances in neural information processing systems},
  volume={37},
  pages={6246--6266},
  year={2024}
}

@inproceedings{shen2025imagdressing,
  title={Imagdressing-v1: Customizable virtual dressing},
  author={Shen, Fei and Jiang, Xin and He, Xin and Ye, Hu and Wang, Cong and Du, Xiaoyu and Li, Zechao and Tang, Jinhui},
  booktitle={Proceedings of the AAAI Conference on Artificial Intelligence},
  volume={39},
  number={7},
  pages={6795--6804},
  year={2025}
}

@article{shen2025long,
  title={Long-Term TalkingFace Generation via Motion-Prior Conditional Diffusion Model},
  author={Shen, Fei and Wang, Cong and Gao, Junyao and Guo, Qin and Dang, Jisheng and Tang, Jinhui and Chua, Tat-Seng},
  journal={arXiv preprint arXiv:2502.09533},
  year={2025}
}

\end{document}